\documentclass[a4paper,conference]{IEEEtran}

\ifCLASSINFOpdf

\else

\fi

\usepackage{graphicx}
\usepackage{amsmath}
\usepackage{amssymb}
\usepackage{booktabs,ragged2e}
\usepackage[flushleft]{threeparttable}

\usepackage{subfigure}
\usepackage{tikz}
\usepackage{pifont}
\usepackage{url}
\usepackage{hyperref}
\usepackage{arydshln}
\hypersetup{
    colorlinks=true,
    linkcolor=blue,
    filecolor=magenta,      
    urlcolor=purple,
    pdftitle={Overleaf Example},
    pdfpagemode=FullScreen,
    }

\begin{document}
%
\title{Video-SwinUNet: Spatio-temporal Deep Learning Framework for VFSS Instance Segmentation \\}

\author{
    \IEEEauthorblockN{Chengxi Zeng\IEEEauthorrefmark{1}, Xinyu Yang\IEEEauthorrefmark{2},
    David Smithard\IEEEauthorrefmark{3}, 
    Majid Mirmehdi\IEEEauthorrefmark{2}, 
    Alberto M Gambaruto\IEEEauthorrefmark{1}, 
    Tilo Burghardt\IEEEauthorrefmark{2}
    }
    \IEEEauthorblockA{\IEEEauthorrefmark{1}Department of Mechanical Engineering, University of Bristol, UK}
    \IEEEauthorblockA{\IEEEauthorrefmark{2}Department of Computer Science, University of Bristol, UK}
    \IEEEauthorblockA{\IEEEauthorrefmark{3}Queen Elizabeth Hospital, Woolwich, UK\vspace{-20pt}}
    
}

\maketitle

\begin{abstract}
This paper presents a deep learning framework for medical video segmentation. Convolution neural network (CNN) and transformer-based methods have achieved great milestones in medical image segmentation tasks due to their incredible semantic feature encoding and global information comprehension abilities. However, most existing approaches ignore a salient aspect of medical video data - the temporal dimension. Our proposed framework explicitly extracts features from neighbouring frames across the temporal dimension and incorporates them with a temporal feature blender, which then tokenises the high-level spatio-temporal feature to form a strong global feature encoded via a Swin Transformer. The final segmentation results are produced via a UNet-like encoder-decoder architecture. Our model outperforms other approaches by a significant margin and improves the segmentation benchmarks on the VFSS2022 dataset, achieving a dice coefficient of 0.8986 and 0.8186 for the two datasets tested. Our studies also show the efficacy of the temporal feature blending scheme and cross-dataset transferability of learned capabilities. Code and models are  available at \url{https://github.com/SimonZeng7108/Video-SwinUNet}.

\end{abstract}

%
\begin{figure*}[h]
\begin{center}
\begin{tabular}{c} \vspace{-15pt}
\includegraphics[height=6.6cm]{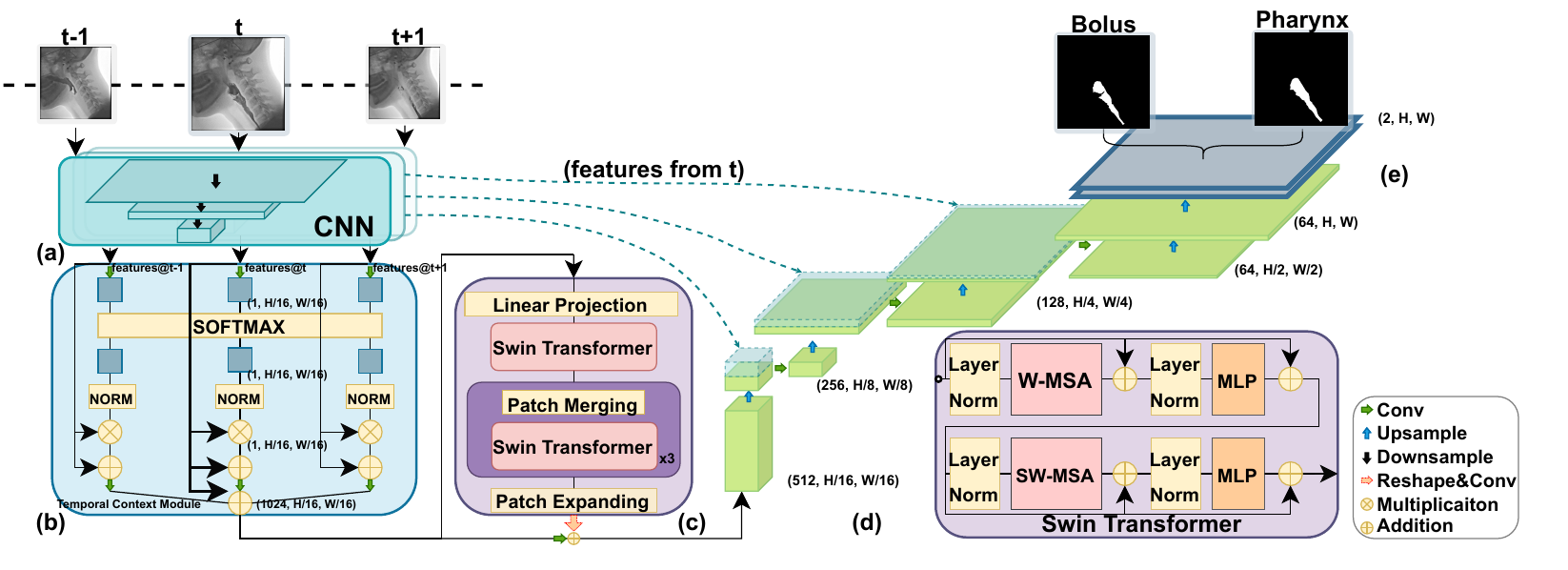} 
\end{tabular}
\end{center}
\caption[example] 
{\label{fig:Video_swin} 
\textbf{Video-SwinUNet Architecture Overview.}(a)A ResNet-50 CNN feature extractor; (b)Temporal Context Module for temporal feature blending; (c)A Swin transformer-based feature encoder; (d)Cascaded CNN up-sampler for segmentation reconstruction; (e)2-layer segmentation head for detailed pixel-wise label mapping. Three skip connections are bridged between the CNN feature extractor and up-sampler as well as from the temporal features.
}
\vspace{-10pt}
\end{figure*} 
\vspace{-10pt}\section{Introduction and Related Work}
Dysphagia or swallowing difficulty is a common complication found in 30 - 50\% of people following stroke~\cite{Smithard2007LongtermOA}. The prevalence of dysphagia in older people with dementia can be high up to 84\%. Risks are identified in people with dysphagia such as malnutrition, development of pneumonia and aspiration. Serious Dysphagia can lead to a strong association with mortality~\cite{Smithard2016DysphagiaAG, Smithard1997TheNH}. Hence early detection and treatment of Dysphagia are crucial.

A Videofluoroscopic Swallow Study (VFSS) is accepted as the gold standard assessment for dysphagia. During VFSS, patients are asked to swallow texture-modified foods and liquids that contain barium. It provides visual data on the trajectory of bolus, muscle, and hyoid bone movement and the connection between anatomy and aspiration ~\cite{Ramsey2003EarlyAO}. However, the clinical assessment requires an extensively experienced speech-language therapist to analyse the visual data on a per-frame basis. The visual data sometimes has both low spatial and temporal quality due to device modalities and radiation noise. Moreover, there can be ambiguity and inconsistency in the judgments of different clinical experts. 


Early attempts at automated processing of the data have used traditional methods, such as Hough transforms~\cite{Zheng2004AutomatedSO}, Sobel Edge detection~\cite{Kellen2009ComputerAssistedAO} and Haar classifiers~\cite{Noorwali2013SemiautomaticTO}
to track lumbar vertebrae, hyoid bone and epiglottis, which are important anatomical structures in the pharyngeal swallowing reflex. With the more recent impact of deep learning in medical image analysis, others have shown advances in pharyngeal phase detection \cite{Lee2018DetectionOT, Lee2019AutomaticDO, Lee2021AutomaticPP, Lee2020MachineLA} and hyoid bone detection \cite{Kim2021HyoidBT, Lee2020OnlineLF, Feng2021AutomaticHB, Iyer2019DeepLA}.

Bolus trajectory is one of the main indicators in a
VFS study, but there are few studies on the automation of bolus detection or segmentation ~\cite{Caliskan2020AutomatedBD, Zhang2021DeepLearningBased, Zeng2022VideoTransUNetTB}. CNN-based works, such as~\cite{Ronneberger2015UNetCN}, demonstrate significant superiority in feature extraction, though they are disadvantaged in computing long-range relations due to their inherent local operations. Vision transformers~\cite{Dosovitskiy2020AnII, Liu2021SwinTH}, on the other hand, have exhibited great predominance in modelling global contextual correlations by using attention mechanisms. 

Recent works that leverage vision transformers~\cite{Chen2021TransUNetTM, Cao2021SwinUnetUP} have shown remarkable performance in medical image segmentation. While others have dealt with video dynamics for detection or segmentation in videos, such as Cao et al. \cite{Cao2019GCNetNN} and Yang et al.~\cite{Yang2019GreatAD},  only few has explicitly addressed the use of temporal information in assisting the detection or segmentation of sequential medical data~\cite{9506463}. The dynamics of bolus suggest that an implicit temporal relationship between the frames on a feature level can be exploited in learning detection or segmentation models. 

In this paper, we present a deep-learning pipeline that takes account of multi-rater annotations and fuses them into a more consistent and reliable ground truth. Subsequently, an architecture is proposed (see Fig.~\ref{fig:Video_swin}) comprising a ResNet-50 feature extractor, a Temporal Context Module (TCM) feature blender, a non-local attention encoder (Swin Transformer) and a cascaded CNN decoder for detailed segmentation map prediction. 

Our main contributions are summarised as follows: i) we provide the VFSS2022 dataset Part 2 in different modalities in contrast to Part 1 annotated with reliable labels for the laryngeal bolus and pharynx. ii) we propose a new architecture enhancing the performance of previous work~\cite{Zeng2022VideoTransUNetTB} by extending the vision transformer encoder to a stronger and more generalised Swin Transformer, iii) we perform a detailed ablation study to reveal the importance of temporal feature blending. We also explore the cross-dataset transferability and generalizability of our deep neural networks on data across different modalities.

\section{Methodology}
\subsection{Architecture Overview}
Inspired by UNet \cite{Ronneberger2015UNetCN}, we follow the encoder-decoder structure to build our video instance segmentation network, as shown in Fig.~\ref{fig:Video_swin}. It takes a video snippet as an input that consists of a sequence of frames with dimension $\mathbf{x} \in \mathbb{R}^{t \times H \times W}$, where $H \times W$ represents the spatial resolution of the input and $t$ is a temporal range of the input sequence.
The input frames will be successively fed into a ResNet-50 backbone for feature extraction (see Fig.~\ref{fig:Video_swin}(a)). Then, the extracted features are simultaneously passed into a novel Temporal Context Module(TCM) (see Fig.~\ref{fig:Video_swin}(b))~\cite{Yang2019GreatAD}, which blends the past and future frame features into the target central frame feature.
Thereafter, the output feature that is integrated with high-level spatial and temporal representation is tokenised into image patches by a Swin transformer encoder (see Fig.~\ref{fig:Video_swin}(c)) for global context construction.
Finally, an up-sampling decoder (see Fig.~\ref{fig:Video_swin}(d)) reconstructs the segmentation map to the original image size of $H \times W$ with cascaded CNNs and binary segmentation heads (see Fig.~\ref{fig:Video_swin}(e)).

\subsection{Temporal Context Module}
The proposed architecture contains a key component Temporal Context Module(TCM) following success in video detection~\cite{Yang2019GreatAD}. The design of TCM follows the principled blending framework by~\cite{Cao2019GCNetNN} where a trainable self-attention module is formulated to a range of frame features from the previous CNN block. The input features $x_t \in\{x_1, \ldots, x_i\}$ are separately linear embedded to a feature space by function $e(\cdot)$ and weights $w_t$ in a concurrent manner. After that,  a global $\mathbf{Softmax}$ operation is applied so that the temporal correlation across all the frames in the feature space can be aggregated.
The agglomerated features are dispersed again to several stems for further linear embeddings. The normalisation of each stem is necessary to prevent vanishing/exploding gradients and can be done easily by $\hat{x}_{t, i}=\frac{1}{H W} \mathcal{C}\left(x_{t, i} ; w_t\right) \sum_{j=1}^{H W} \mathcal{C}\left(x_{t, j} ; w_t\right)$. Identity mapping operations by $\mathbf{multiplication} \otimes$ and $\mathbf{addition} \oplus$ are applied in each stem. In the end, stabilised features are added back to the central frame feature as a final single mixture high-level description of the short-term snippet. In summary, the TCM operation can be formulated by:
\begin{equation}
z_{t, i}^{T C M}=x_{t, i}+\sum_{n \in T} w_{n}^{**}\left(x_{n, i} \oplus w_{n}^{*} \sum_{j=1}^{HW} \hat{x}_{n, j} \otimes x_{n, j}\right)~,
\end{equation}\\
where $x_n$ are the linear embedded features to be combined, $w_n^{*}$ and $w_n^{**}$ are trainable parameters for identity additions and blending operations.

\begin{figure*}
\begin{center}
\begin{tabular}{c} \vspace{-10pt}\hspace*{-0.4cm}
\includegraphics[height=3.5cm]{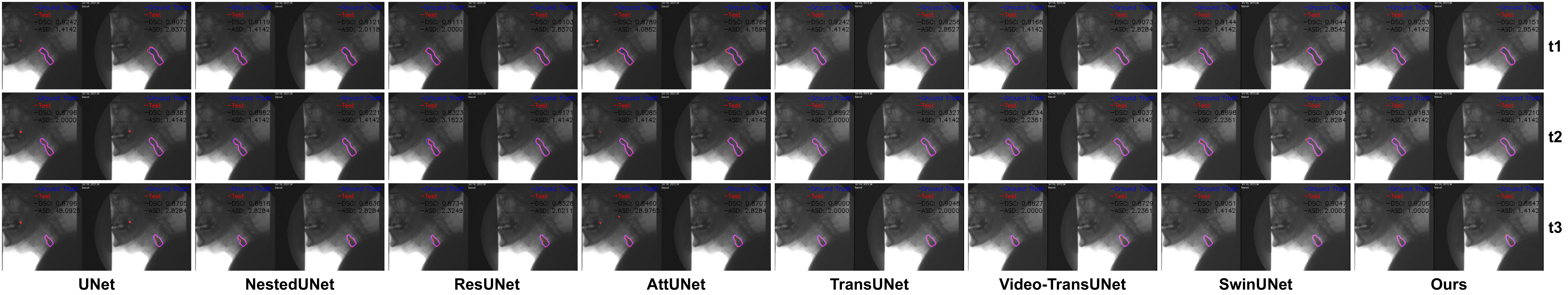}
\end{tabular}
\end{center}
\caption[example] 
{\label{fig:qualitative} 
\textbf{Qualitative Results.} Model segmentation results on 3 consecutive frames selected from VFSS Part2 dataset testset. All results are in instance pairs of bolus and pharynx predictions side by side. The red and blue outlines indicate the output segmentation and ground truth, respectively.(Best viewed zoomed)}
\vspace{-10pt}
\end{figure*} 
\subsection{Swin Transformer} 
Following~\cite{Dosovitskiy2020AnII},  we tokenise temporal blended features into feature patches $x_p$ and map them into a latent D-dimensional embedding space via learnable linear projection. \cite{Liu2021SwinTH, Cao2021SwinUnetUP} suggest the unnecessity of employing position embedding $\mathbf{E_{pos}}$ in Swin transformer, hence we omitted it in our work for simplicity. The projected feature can be expressed as $z_{0}=\left[x_{p}^{1} \mathbf{E_{pat}} ; \cdots ; x_{p}^{N} \mathbf{E_{pat}}\right]$, where $\mathbf{E_{pat}}$ is the patch embedding linear projection.
The conventional vision transformer computes global attention across the vectorized patches, As a result, the computational complexity is quadratically increased along with the increase of the input resolutions. To alleviate the computation overhead in Multihead Self-Attention($\mathbf{MSA}$), a Window-based MSA($\mathbf{W\mbox{-}MSA}$) method is proposed in~\cite{Liu2021SwinTH}. The window moves along the feature or image without overlapping and conducts local self-attention. The sizes of the windows vary from 4 to 16 patches  through deeper layers. Such hierarchical feature representation makes it more computationally efficient. The $\mathbf{W\mbox{-}MSA}$ can be expressed as:
\begin{equation}
\operatorname{Attention}(Q, K, V)=\operatorname{SoftMax}\left(Q K^T / \sqrt{d}+B\right)V~,
\end{equation}
where $Q, K, V \in \mathbf{R}^{M^2 \times d}$ stands for the query, key and value matrices respectively; $d$ is the query/key dimension, and $M^2$ patch numbers in a window. The relative position biases $B$ , are taken from the bias matrix $\hat{B}$.

To model the relationship between windows, a Shifted-Window MSA ($\mathbf{SW\mbox{-}MSA}$) is proposed in~\cite{Liu2021SwinTH}, the patches take turns in  two consecutive Swin Transformer blocks, each of which contains both a $\mathbf{W\mbox{-}MSA}$ and a $\mathbf{SW\mbox{-}MSA}$ accompanied with a 2-layer $\mathbf{MLP}$ followed a $\mathbf{GELU}$ activation function. And $\mathbf{LayerNorm}$(LN) and skip connections are added before the $\mathbf{MLP}$, as illustrated in Fig.~\ref{fig:Video_swin}.

\begin{table*}[ht]
\caption{\textbf{Quantitative Results.} Segmentation accuracy on 5 metrics of VFSS2022 Part1/Part2 is shown, as well as a number of parameters and FLOPs of each model. Part1/Part2 datasets are trained separately.} 
\vspace{-10pt}
\label{tab:comparison method}
\begin{center}       
\begin{footnotesize}    
\begin{tabular}{l|l|l|l|l|l|l|l} 
\toprule
\rule[-1ex]{0pt}{3.5ex}  \textbf{Model} & \textbf{DSC} & \textbf{HD95} & \textbf{ASD} & \textbf{Sensitivity} & \textbf{Specificity} & \textbf{FLOPs} & \textbf{\#Params}\\
\midrule \midrule
\rule[-1ex]{0pt}{3.5ex}  (1) UNet\cite{Ronneberger2015UNetCN}  & $0.8422/0.7894$ & $14.7530/20.7516$ & $2.1675/4.6458$ & $0.8289/0.7414$ & $0.9988/0.9793$ & 50.1G & 34.5M\\

 \rule[-1ex]{0pt}{3.5ex}  (2) NestedUNet\cite{Stoyanov2018DeepLI}  & $0.8335/0.7537$ & $13.7601/6.4952$ & $2.2275/5.1220$ & $0.8305/0.7188$ & $0.9987/0.9682$ & 105.7G &36.6M\\

\rule[-1ex]{0pt}{3.5ex}  (3) ResUNet\cite{Zhang2017RoadEB} & $0.8465/0.7846$ & $11.982/6.4187$ & $2.0487/2.4218$ & $0.8183/0.7218$ & $\mathbf{0.9991}/0.9994$ & 43.1G & 31.5M \\

\rule[-1ex]{0pt}{3.5ex}  (4) AttUNet\cite{Oktay2018AttentionUL}  & $0.8501/0.7917$ & $12.9356/16.9552$ & $2.1832/4.2174$ & $0.8328/0.7721$ & $0.9988/0.9985$  & 51.0G & 34.8M\\

\rule[-1ex]{0pt}{3.5ex} (5)  TransUNet\cite{Chen2021TransUNetTM}  & $0.8586/0.8046$ & $7.4510/4.6291$ & $1.1050/1.9322$ & $0.8486/0.7579$ & $0.9989/0.9929$ & 29.3G &105.3M\\

\rule[-1ex]{0pt}{3.5ex}  (6) Video-TransUNet\cite{Zeng2022VideoTransUNetTB}  & $0.8796/0.8041$ & $6.9155/4.7775$ & $\mathbf{1.0379}/1.5270$ & $0.8851/0.7423$ & $0.9986/\mathbf{0.9996}$ & 40.4G & 110.5M \\

\rule[-1ex]{0pt}{3.5ex}  (7) SwinUNet\cite{Cao2021SwinUnetUP}  & $ 0.8477/0.8001$ & $10.2897/5.9846$ & $2.0817/2.1342$ & $0.8459/0.7336$ & $0.9985/0.9935$ & 6.1G &27.1M \\

\rule[-1ex]{0pt}{3.5ex}  (8) Video-SwinUNet(ours)  & $\textbf{0.8986}/\textbf{0.8186}$ & $\textbf{6.2365}/\textbf{4.5268}$ & $1.3081/\textbf{1.2052}$ & $\textbf{0.9011}/\textbf{0.7756}$ & $0.9986/0.9995$ & \textbf{25.8G}& \textbf{48.9M}\\
\midrule
\end{tabular}
\end{footnotesize}    
\end{center}
\vspace{-20pt}
\end{table*}

\section{EXPERIMENTS AND RESULTS}
\subsection{Datasets and Implementation details}
\textbf{Datasets.} The VFSS2022 datasets are collected in two major hospitals and the utilisation of the anonymised data is ethically reviewed and approved by the hospitals and our internal institutional Ethics Board. During the VFS studies, the patients carried out modified barium swallow tests under the practitioner's supervision. VFSS2022 Part 1 produces 3.5 minutes of swallowing videos which result in 440 sampled frames with a spatial resolution of $512\times 512$ pixels. Each frame is annotated by 3 experts and reviewed by 2 speech and language therapists and compromising labels for bolus and pharynx. The final ground truth is fused together with the 3 labels by a common image fusion strategy STAPLE~\cite{Warfield2004SimultaneousTA}. VFSS2022 Part 2 is annotated by one trained expert consisting of 154 frames and corresponding labels, it appears to have more modal noises and poorer temporal quality, and is used for the model generalisation test.

\textbf{Implementation details.} The bolus and pharynx are concatenated as 2-layer tensors for the end-to-end model co-learning from both. The layers for the frames with no visible bolus are replaced with full-size zeroed tensors. To study the effect of input snippet lengths in our system, the input number of frames is in the range of $t = 3, 5, 7, 9, 11\& 13$ both for training and testing. All experiments supported online data augmentation such as random limited rotation and flipping. We initialise the weights of the ResNet-50 backbone and Swin-Transformer from the pre-trained models~\cite{Chen2021TransUNetTM, Cao2021SwinUnetUP}. During training, our system takes in a batch size of 2 and is equipped with an Adam optimizer with an initial learning rate of 1e-3. For transfer learning, the learning rate is dropped to 1e-4 at the beginning. A learning rate scheduler is set to drop the learning rate to 80\% after 20 epochs of validation loss saturation. The architecture is achieved in Python 3.8.5 and Pytorch 1.9 and trained with an NVIDIA Tesla P100 16GB GPU. We consider the overall loss of Binary Cross Entropy Loss and Dice Loss as the final training objectives.

\subsection{Comparison with the state of the art}

\begin{figure}
\begin{center}
\begin{tabular}{c} \vspace{-14pt}\hspace{-0.8cm}
\includegraphics[height=4.8cm]{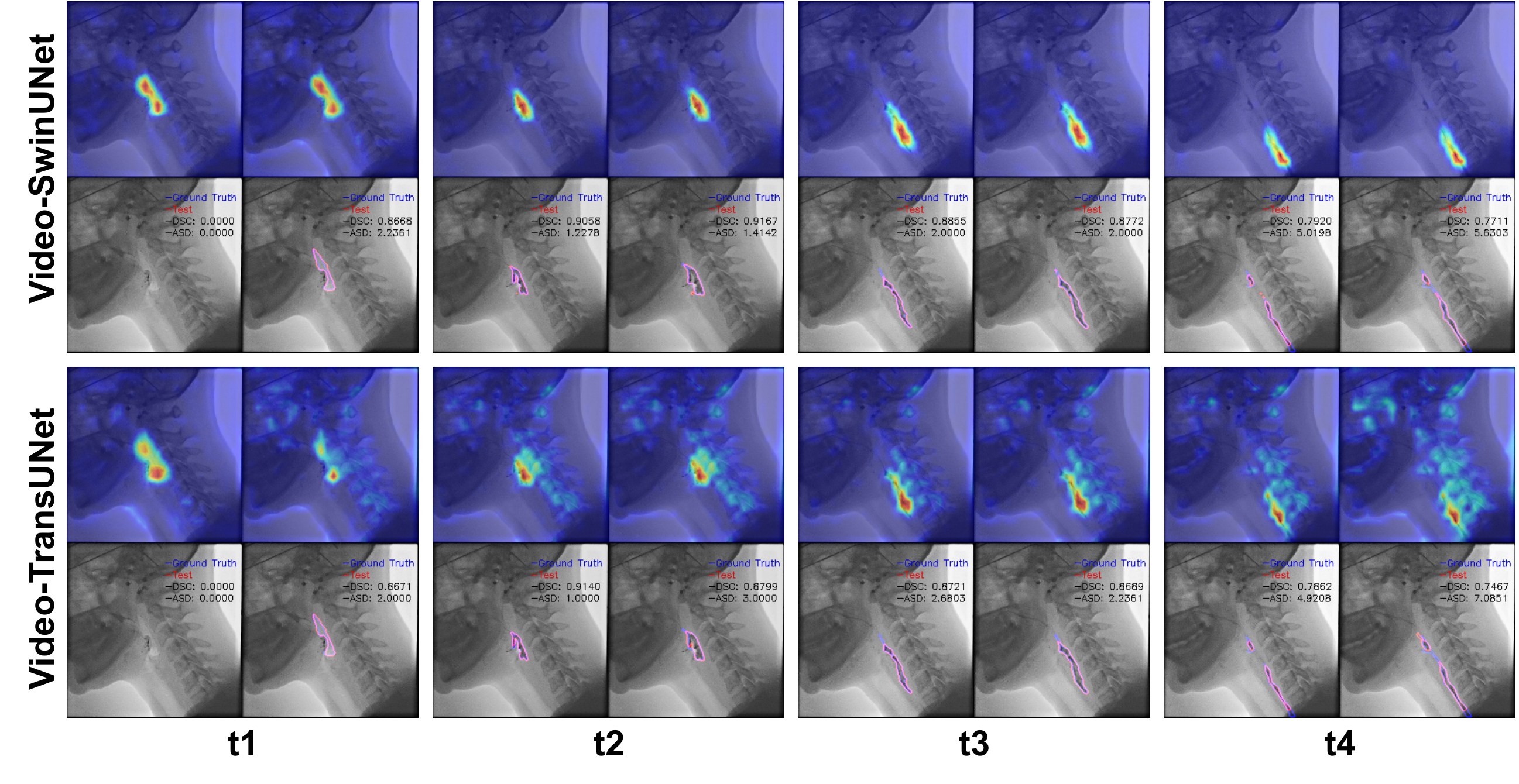}
\end{tabular}
\end{center}
\caption[example] 
{\label{fig:grad_cam} 
\textbf{Grad-CAM Visualisation.} Comparing the two closest competing architectures, grad-cam maps show where the model pays attention. Note the cleaner focus of our proposed approach.(Best viewed zoomed)}
\vspace{-14pt}
\end{figure}

We compare our proposed architecture with major medical image segmentation models including UNet~\cite{Ronneberger2015UNetCN}, NestedUNet~\cite{Stoyanov2018DeepLI}, ResUNet~\cite{Zhang2017RoadEB}, AttUNet~\cite{Oktay2018AttentionUL}, TransUNet~\cite{Chen2021TransUNetTM}, Video-TransUNet~\cite{Zeng2022VideoTransUNetTB} and SwinUNet~\cite{Cao2021SwinUnetUP} over 5 common evaluation metrics, the Dice Coefficient~(DSC), the 95th percentile of the Hausdorff Distance~(HD95), the Average Surface Distance~(ASD), Sensitivity and Specificity, see Tab.~\ref{tab:comparison method}. Additionally, we also include the total number of parameters of the model and the total floating-point operations(FLOPs) to compare the model size and computing performance. It can be seen that in Tab.~\ref{tab:comparison method}, our method improved segmentation accuracy to $89.86\%$/$81.86\%$(DSC) and $6.2365$/$4.5268$ pixels(HD95) on VFSS2022 Part1/Part2, the test results dominate the previous SOTA~\cite{Zeng2022VideoTransUNetTB} and other methods with a significant margin.
The general quality is greatly improved and output noises are less produced, as demonstrated in the qualitative results, see Fig.~\ref{fig:qualitative}. More importantly, the proposed method achieves a remarkable speed-accuracy trade-off. Although compared with SwinUNet~\cite{Cao2021SwinUnetUP} the model size is doubled, remaining as the second least parameterised model in comparative study, it notably improved the segmentation accuracy by 5.09\%. it is also observed that our model has reduced the number of parameters to less than half compared with the previous SOTA while not sacrificing computational efficiency due to the design of hierarchical shifting windows.  Fig.~\ref{fig:grad_cam} shows the Grad-CAM output one layer before TCM in Video-TransUNet and Video-SwinUNet. The attention maps from our method devote great concentrations are computed to task-relevant features. Hence it promotes the efficacy of the TCM and the importance of temporal-relation constructions.

\subsection{Ablation study}

\begin{table}[b]
\vspace{-15pt}
\caption{Ablation Study on impacts of different encoder-decoder combinations to performances.} 
\vspace{-15pt}
\label{tab:ablation_parts}
\begin{center}       
\begin{tabular}{l l|l|l|l} 
\toprule
\rule[-1ex]{0pt}{3.5ex}  \textbf{Encoder} & \textbf{Decoder} & \textbf{DSC} & \textbf{HD95} & \textbf{\#Params} \\
\midrule \midrule
\rule[-1ex]{0pt}{3.5ex} (1)Swin & Swin & $0.8477$ & $10.2897$ & 27.1M\\

 \rule[-1ex]{0pt}{3.5ex} (2)CNN+Swin & Swin & $0.8483$ & $9.5757$ & 39.1M\\

\rule[-1ex]{0pt}{3.5ex} (3)CNN+Swin &  Swin+CUP & $0.8562$ & $8.0544$ & 43.8M\\

\rule[-1ex]{0pt}{3.5ex} (4)CNN+TCM + Swin & Swin+CUP& 
 $0.8634$ & $6.8941$ & 49.1M \\

\rule[-1ex]{0pt}{3.5ex} (5)CNN+Swin & CUP & $0.8592$ & $8.2744$ & 43.2M\\

\rule[-1ex]{0pt}{3.5ex} (6)CNN+TCM+Swin & CUP & $0.8899$ & $\mathbf{5.1234}$ & 48.4M\\

\rule[-1ex]{0pt}{3.5ex} (7)S/A+TSC & S/A & $\mathbf{0.8986}$ & $6.2365$ & 48.9M\\
\midrule
\end{tabular}
\end{center}
\vspace{-20pt}
\end{table}
We conducted major ablation experiments to reveal the efficacy of the proposed temporal blending framework via a novel TCM component. We modulate 4 main components, CNN extractor(CNN), Swin Transformer Block(Swin), Temporal Context Module(TCM) and CNN up-sampler(CUP) in our experiments. Comparing Tab.~\ref{tab:ablation_parts}(4) to (3) and (6) to (5), we can see that TCM has increased performance by a margin without an extra cost in computational power. The CNN feature extractor and CUP indicate the effectiveness of convolutional operations due to their intrinsic locality characteristic. The use of skip connections is studied in ~\cite{Ronneberger2015UNetCN, Chen2021TransUNetTM}, we attached an additional Temporal feature Skip Connection(TSC) to the decoder path, see Tab.~\ref{tab:ablation_parts}(7)(Video-SwinUNet), it is suggested that the TSC is beneficial in constructing the segmentation map, which further supports the significance of temporal features in the neural network. Grid search over snippet sizes $t = 3, 5, 7, 9, 11 \& 13$ revealed the optimal, application-specific size $t = 5$ both for training and testing.

\subsection{Transfer learning}
\begin{table}[b]
\vspace{-15pt}
\caption{Transferability test on each part of the model.} 
\vspace{-15pt}
\label{tab:transfer_learning}
\begin{center}       
\begin{tabular}{l|l|l|l|l|l} 
\toprule
\rule[-1ex]{0pt}{3.5ex}\textbf{Pretrained} &\textbf{Training}&\textbf{Frozen}&\textbf{Fine-tuning}&\textbf{DSC} &\textbf{HD95}\\ \textbf{dataset} & \textbf{dataset} & \textbf{weights} & \textbf{weights} &  \\
\midrule \midrule
\rule[-1ex]{0pt}{3.5ex} (1)Part1 & Part2 & N/A & All & 0.7618 & 15.1496 \\

\rule[-1ex]{0pt}{3.5ex} (2)Part1 & Part2 & a & b+c+d+e & 0.7979 & 4.9123\\

\rule[-1ex]{0pt}{3.5ex} (3)Part1 & Part2 & a* & b+c+d+e & \textbf{0.8437} & \textbf{4.6512}\\

\rule[-1ex]{0pt}{3.5ex} (4)Part1 & Part2 & a + b & c+d+e & 0.7295 & 16.4245\\

\rule[-1ex]{0pt}{3.5ex} (5)Part1 & Part2 & a+b+c & d+e & 0.7030 & 18.1302\\

\rule[-1ex]{0pt}{3.5ex} (6)Part1 & Part2 & a*+b+c* & d+e & 0.8171 & 5.2039\\ \hdashline

\rule[-1ex]{0pt}{3.5ex} (7)Part2 & Part1 & a* & b+c+d+e & \textbf{0.8920} & \textbf{4.1984}\\

\rule[-1ex]{0pt}{3.5ex} (8)Part2 & Part1 & a* & b+c*+d+e & 0.8850 & 6.0604\\ \hdashline

\rule[-1ex]{0pt}{3.5ex} (9)Part1+2 & Part1 & a* & b+c*+d+e & \textbf{0.8994} & \textbf{3.8415}\\

\rule[-1ex]{0pt}{3.5ex} (10)Part1+2 & Part2 & a* & b+c*+d+e & 0.8094 & 4.6473\\
\midrule
\end{tabular}
\end{center}
\vspace{-20pt}
\end{table}

We also explore the transferability of each component in Fig.~\ref{fig:Video_swin}, CNN(a), TCM(b), Swin Transformer(c), Decoder(d), Segmentation head(e), shown in Tab.~\ref{tab:transfer_learning}, a* indicates weights are pre-trained on ImageNet from~\cite{Chen2021TransUNetTM}, otherwise are trained from scratch. We adopt a standard transfer learning approach, fine-tuning, to investigate the generalisation ability of each part in domain shift from VFSS2022 Part 1 to Part 2 and vice versa. It is suggested that fine-tuning the later part after feature extraction is beneficial in domain adaption in both ways, noting row(3) and row(7). It is also shown our model's ability to generalise in part 1/part 2 when the model trains the entire dataset and can even gain performance boosts(DSC 89.94\%) in part 1, see row(10) and row(11).

\section{Conclusion}
We presented an end-to-end framework that exploits multi-frame inputs to segment VFSS2022 data with great success leading to performance gains and model size reduction. Our proposed neural network merits local and global spatial context and leveraged temporal features. Each of the modules can be fine-tuned or exchanged. The final framework achieves superior performance over other designs and provides a new, alternative pipeline for medical video segmentation tasks. \ \\

\begin{footnotesize}
ACKNOWLEDGEMENTS. Data usage and publication are granted by UoB Ethics Approval REF: 11277. We thank project investigators Ian Swaine, Salma Ayis, Aoife Stone-Ghariani, Dharinee Hansjee, Stefan T Kulnik, Peter Kyberd, Elizabeth Lloyd-Dehler, William Oliff, Lydia Morgan and Russel Walker and thank Yuri Lewyckyj and Victor Perez for their annotations. Project: CTAR-SwiFt; Funder: NIHR; Grant: PB-PG-1217-20005.
\end{footnotesize}

\bibliographystyle{IEEEtran}
\bibliography{IEEEexample}

\end{document}